\documentclass[sigconf]{acmart}

\usepackage{multirow}
\usepackage{xcolor}
\usepackage{graphicx} 
\usepackage{booktabs} 
\usepackage{graphicx}
\usepackage{amsmath}
\usepackage{amsthm}
\usepackage{textcomp}
\usepackage{algorithm}
\usepackage{algorithmic}

\usepackage[switch]{lineno}
\usepackage{soul}
\usepackage{url}
\usepackage{bm}      

\copyrightyear{2025}
\acmYear{2025}
\setcopyright{acmlicensed}
\acmConference[MM '25] {Proceedings of the 33rd ACM International Conference on Multimedia}{October 27--31, 2025}{Dublin, Ireland.}
\acmBooktitle{Proceedings of the 33rd ACM International Conference on Multimedia (MM '25), October 27--31, 2025, Dublin, Ireland}
\acmISBN{979-8-4007-2035-2/2025/10}
\acmDOI{10.1145/3746027.3755751}

\begin{document}

\title{AlignCAT: Visual-Linguistic Alignment of Category and Attribute for Weakly Supervised Visual Grounding}


\author{Yidan Wang}
\authornote{Equal contribution.} 

\orcid{0009-0007-9439-5291}
\affiliation{%
  \institution{Nanjing University of Aeronautics and Astronautics}
  \city{Nanjing}
  \country{China}
}
\email{wangyidan@nuaa.edu.cn}

\author{Chenyi Zhuang}
\authornotemark[1] 
\orcid{0009-0004-0700-5645}
\affiliation{%
  \institution{University of Trento}
  \city{Trento}
  \country{Italy}
}
\email{chenyi.zhuang@unitn.it} 

\author{Wutao Liu}
\orcid{0009-0009-5376-5289}
\affiliation{%
  \institution{Nanjing University of Aeronautics and Astronautics}
  \city{Nanjing}
  \country{China}
}
\email{wutaoliu@nuaa.edu.cn} 

\author{Pan Gao}
\authornote{Corresponding author}
\orcid{0000-0002-4492-5430}
\affiliation{%
  \institution{Nanjing University of Aeronautics and Astronautics}
  \city{Nanjing}
  \country{China}
}
\email{pan.gao@nuaa.edu.cn}

\author{Nicu Sebe}
\orcid{0000-0002-6597-7248}
\affiliation{%
  \institution{University of Trento}
  \city{Trento}
  \country{Italy}
}
\email{niculae.sebe@unitn.it} 

\renewcommand{\shortauthors}{Wang et al.}

\begin{abstract}
Weakly supervised visual grounding (VG) aims to locate objects in images based on text descriptions. Despite significant progress, existing methods lack strong cross-modal reasoning to distinguish subtle semantic differences in text expressions due to category-based and attribute-based ambiguity. To address these challenges, we introduce AlignCAT, a novel query-based semantic matching framework for weakly supervised VG. To enhance visual-linguistic alignment, we propose a coarse-grained alignment module that utilizes category information and global context, effectively mitigating interference from category-inconsistent objects. Subsequently, a fine-grained alignment module leverages descriptive information and captures word-level text features to achieve attribute consistency. By exploiting linguistic cues to their fullest extent, our proposed AlignCAT progressively filters out misaligned visual queries and enhances contrastive learning efficiency. Extensive experiments on three VG benchmarks, namely RefCOCO, RefCOCO+, and RefCOCOg, verify the superiority of AlignCAT against existing weakly supervised methods on two VG tasks. Our code is available at: \href{https://github.com/I2-Multimedia-Lab/AlignCAT}{https://github.com/I2-Multimedia-Lab/AlignCAT}.
\end{abstract}

\ccsdesc[500]{Computing methodologies~Image segmentation; Scene understanding}

\keywords{Weakly Supervised Visual Grounding, Multimodality}

\maketitle

\section{Introduction}
\label{Sec:introduction}
Visual Grounding (VG) aims to identify objects in an image corresponding to a given text description and has gained attention for its potential in open-ended detection for various computer vision applications~\cite{stefanini2022show,marino2019ok,yu2019deep}. While fully supervised methods~\cite{zhu2022seqtr,yang2023improving,dai2024simvg,luo2020multi,zhou2021real} achieve high accuracy, they rely on instance-level annotations, which are both labor-intensive and time-consuming to obtain. To alleviate this burden, recent studies have explored weakly supervised learning in two grounding tasks, namely Referring Expression Comprehension (REC) and Referring Expression Segmentation (RES). These works have reframed weakly supervised VG as a region-based~\cite{DTWREG,IGN,eiras2024segment}, anchor-based~\cite{Refclip,APL}, or query-based~\cite{QueryMatch} matching problem. However, understanding complex textual descriptions and associating referents in multi-object images remains challenging.


\begin{figure*}[t!]
	\centering
	\includegraphics[width=0.95\textwidth]{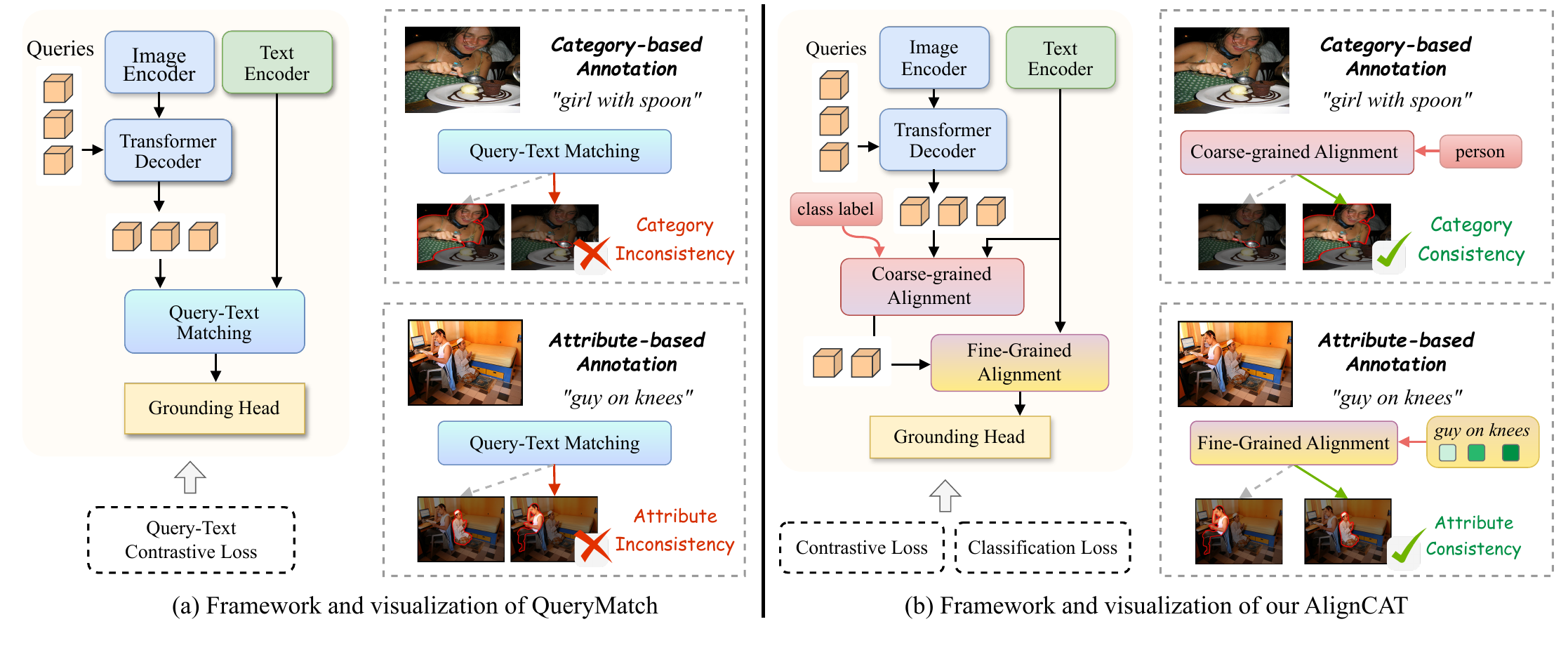}
    \vspace{-7mm}
	\caption{Comparison of QueryMatch and the proposed AlignCAT. (a) QueryMatch fails to deal with category-based and attribute-based ambiguity in annotations. (b) AlignCAT progressively leverages linguistic cues from coarse (right-top) to fine (right-bottom) to filter visual queries, achieving category and attribute consistency.}
    \vspace{-2mm}
	\label{fig:framework_comparison}
\end{figure*}

In Figure \ref{fig:framework_comparison}, we identify two types of text annotation in existing VG benchmarks: \textbf{(1) category-based annotations} that distinguish the referred object in its fundamental class from objects in other categories. For example, in the sentence \emph{``girl with spoon''}, two objects \emph{``girl''} and \emph{``spoon''} belong to different classes. The model should identify the logical object \emph{``girl''} rather than the contextual object \emph{``spoon''}, and align this linguistic category information with visual features. \textbf{(2) attribute-based annotations} that describe specific characteristics of the referred object, such as colors and spatial relations. For example, the sentence \emph{``guy on knees''} has no conflict in the person category, but its descriptive information \emph{``on knees''} poses a challenge for the model in identifying the referred object as the image contains multiple persons. This requires an understanding of nuanced textual and visual semantics. However, the state-of-the-art query-based method \cite{QueryMatch} fails to produce reliable grounding results on both annotation types. While this method utilizes contrastive learning to amplify the alignment of target texts and positive queries, it is not conducive to discriminating nuanced semantics in object categories and attributes. For the category-based annotation, the contextual object \emph{``spoon''} is used to enrich the information of the target object \emph{``girl''}, yet it has created activation noise and hindered the accurate visual-linguistic alignment, showing category inconsistency. For the attribute-based annotation, it mismatches the visual features of the incorrect guy to the target action \emph{``on knee''}, showing attribute inconsistency. 



To address the aforementioned visual-linguistic inconsistencies, we propose \textbf{AlignCAT} (\textbf{Align} \textbf{C}ategory then \textbf{AT}tribute), a novel query-based VG framework. To ensure category consistency, we first design a \textbf{coarse-grained alignment} module that leverages category information and the global context from the input textual expression. This coarse alignment mitigates interference from irrelevant objects, effectively narrowing down the search space for ideal visual queries. To achieve attribute consistency, we further propose a \textbf{fine-grained alignment} module that employs adaptive phrase attention to capture word-level descriptive linguistic features. Such a finer alignment enhances cross-modal correspondences and resolves intra-class ambiguities when multiple visual objects belong to the same category.  By aligning visual queries first at a coarse level and then at a finer level, AlignCAT highlights the key role of linguistic cues in understanding cross-modal representations. This category-then-attribute progressive alignment within a contrastive learning framework significantly enhances VG performance. In summary, the main contributions of this work are three-fold: 
\begin{itemize}
    \item We identify category inconsistency and attribute inconsistency in existing weakly supervised VG methods. To address these challenges, we propose a novel query-based category-then-attribute matching framework, modeling linguistic representations from general to detailed levels.
    \item To achieve visual-linguistic alignment, we design a coarse-grained module that leverages category information and global context to filter out category-inconsistent visual queries, and a fine-grained module that employs adaptive phrase attention to ensure attribute consistency.
    \item Evaluated on three benchmarks of REC and RES, our proposed method achieves state-of-the-art performance, demonstrating the potential of linguistic cues and the efficacy of the category-then-attribute matching strategy in enhancing visual-linguistic alignment.
\end{itemize}

\section{Related Works}

\textbf{Referring Expression Segmentation (RES)}. This task aims to overcome the efficiency limitation of fully supervised learning schemes. Weakly supervised RES does not require intensive pixel-level annotation, which is less expensive and more efficient for training. Several works \cite{TSEG,kim2023shatter} achieve region-text matching through multi-instance learning, but are far inferior to fully supervised methods. 
Instead of aggregating visual entities, TRIS \cite{TRIS} extracts rough object locations as pseudo-labels based on the input text to perform object localization. Lee et al. \cite{lee2023weakly} relies on the linguistic relationship, which predicts significant maps for each word. However, the masks generated by these methods are highly noisy, resulting in less accurate segmentation.

\begin{figure*}[htp]
	\centering
	\includegraphics[width=1\textwidth]{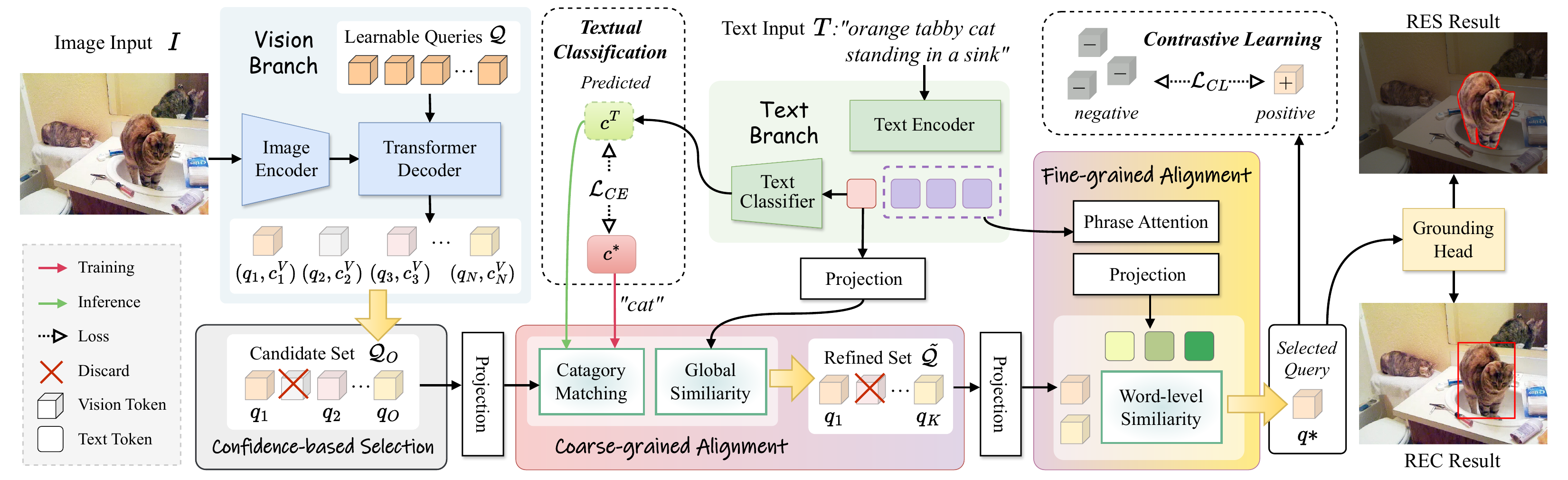} %
    \vspace{-6mm}
	\caption{AlignCAT framework overview. AlignCAT filters visual queries by hierarchically leveraging linguistic cues. The coarse-grained alignment module utilizes category and global information to discard category-inconsistent candidates. The fine-grained alignment module employs adaptive phrase attention to select the attribute-consistent visual query.}
	\label{fig:AlignCAT_framework}
\end{figure*}

\noindent\textbf{Referring Expression Comprehension (REC)}. Compared to fully supervised REC, weakly supervised REC is more challenging due to the lack of bounding box annotations. To obtain additional supervision signals, existing REC methods \cite{EARN,wang2019phrase} incorporate external knowledge and align the region-based information with the corresponding phrases. Some works~\cite{chen2018knowledge,liu2019knowledge} further utilize prior knowledge to filter out irrelevant region proposals. Recent advances also include leveraging language models to build negative samples~\cite{zhang2023cycle}, or pre-trained models to generate pseudo-labels~\cite{liu2023confidence,jiang2022pseudo}. Yet, these two-stage methods lack generalization to real-world scenarios and large-scale tasks. To improve efficiency, anchor-based methods \cite{Refclip,APL} remove the region proposal stage towards a one-stage process. QueryMatch \cite{QueryMatch} further introduces a query-text matching scheme to improve the learning of object representations. Despite their advances in efficiency, we identify the challenges of category and attribute inconsistency in existing one-stage methods. To address this problem, our method leverages linguistic cues from general to specific, integrating category information and global context for coarse-grained alignment, and then exploiting word-level descriptions for fine-grained alignment. The category-then-attribute matching framework significantly improves VG results for both RES and REC tasks, especially in multi-object scenarios with complex text expressions.

\section{Preliminary}
Following~\cite{QueryMatch}, we reformulate VG as a query-text matching problem by adopting a query-based detector Mask2Former \cite{mask2former}. It establishes one-to-one associations with objects in the image by $N$ learnable vectors, namely \textit{queries}, denoted as $\mathcal{Q}=\{q_1, ..., q_N\}$. QueryMatch filters out noisy and low-quality query features based on their confidence scores, resulting in a candidate set $\mathcal{Q}_O$, where $O$ is a pre-defined hyperparameter. This method defines two metrics as \textit{difficulty} and \textit{uniqueness} to quantitatively estimate the quality of negative samples. Specifically, difficulty measures vision-language alignment, while uniqueness requires that high-quality negative queries significantly differ from other queries in the embedding space. Given a set of candidate queries, QueryMatch iteratively estimates the quality of $i$-th query as:
\begin{equation}
\label{eq:negative_difficulty}
    S_{d_i} = Norm(\text{sim}(f_{q_i}, f_{t})),
\end{equation}
\begin{equation}
\label{eq:negative_uniqueness}
    S_{u_i} = Norm(-\max_{j=1}^M \cos(f_{q_i}, f_{q_j})),
\end{equation}
where $f_{q_i}$ is the feature of the current negative query, $f_{q_j}$ is the feature of a previously selected negative query, $f_{t}$ is the text feature. $S_{d_i}$ is the difficulty of the $i$-th query, measured by the dot product similarity between visual and linguistic features, denoted $\text{sim}(f_{q_i}, f_{t})$. $S_{u_i}$ is the uniqueness of the $i$-th query, measured by the cosine similarity between two visual queries, denoted $\cos(f_{q_i}, f_{q_j})$. $Norm(\cdot)$ is the min-max normalization. 
The overall quality score of the negative query is defined as:
\begin{equation}
\label{eq:negative_quality}
    S_{q_i} = S_{d_i} \cdotp S_{u_i}.
\end{equation}
Ranked in descending order, an appropriate number of negative samples is selected to perform contrastive learning.

The effectiveness of QueryMatch relies on precise query-text matching. This method introduces an effective negative query selection scheme, while simply select the positive query by computing the similarity to the global text feature $f_t$. However, textual descriptions are expressive and require strong reasoning abilities to understand them. As presented in Figure \ref{fig:framework_comparison}, QueryMatch fails at identifying visual queries from the candidate set $\mathcal{Q}_O$ to achieve visual-linguistic consistency at the category and attribute levels. In this study, we leverage linguistic cues to their fullest extent, emphasizing both coarse and fine-grained information. Based on the category-then-attribute alignment strategy, our proposed AlignCAT can select high-quality positive visual queries and enhance query-text matching accuracy.

\section{Methodology}

\subsection{Overview of AlignCAT}
Given the input image $I$ and the input expression $T$, we aim to locate the referred object through a bounding box (for REC) or a mask (for RES). To address the challenges of category and attribute inconsistencies, we introduce AlignCAT, a novel query-based weakly supervised VG framework. As illustrated in Figure \ref{fig:AlignCAT_framework}, the main goal of AlignCAT is to select high-quality positive queries through a category-then-attribute matching mechanism for efficient contrastive learning. We follow \cite{QueryMatch} and adopt a query-based detector \cite{mask2former} to process the input image $I$. The encoded image features are then fed into the Transformer decoder to interact with $N$ learnable queries, outputting the query features $\{f_{q_1}, ..., f_{q_N}\}\in \mathbb{R}^{d_v}$ for visual queries in $\mathcal{Q}$, where $d_v$ is the visual dimension, and one-to-one classifications $\{c^V_{1}, ..., c^V_{N}\}$ predicted by the visual classifier.

Unlike existing query-based VG frameworks \cite{QueryMatch}, we leverage linguistic cues in the input expression $T$ to achieve visual-linguistic alignment. The text encoder transforms $T$ into the global feature $f_t \in \mathbb{R}^{d_t} $ and the word-level features $\mathcal{F}_w \in \mathbb{R}^{l\times d_t}$, where $l$ is the length of the input text and $d_t$ is the dimension of the linguistic feature. Our method progressively filters out visual queries through three sequential selection modules: (1) a confidence-based filtering stage reduces the number of visual queries from $N$ to $O$, forming a candidate subset $\mathcal{Q}_O$; (2) a coarse-grained alignment module evaluates category consistency and global query-text similarity to further refine the candidates into a refined query set $\widetilde{Q}$; (3) a fine-grained alignment captures attribute details by recalibrating the word-level features $\mathcal{F}_w$, ultimately selecting the most relevant query ${q}^{*}$. Finally, we can decode the selected visual query to obtain the bounding box or the mask of the referred object through a grounding head:
\begin{equation}
    r^* = \text{Head}({q}^{*}).
\end{equation}

\begin{table*}[t]
    \small
    \caption{Comparisons with state-of-the-art methods on three RES benchmark datasets. Best in \textcolor{red}{red} and second in \textcolor{blue}{blue}.}
    \vspace{-4mm}
    \label{tab:res_results}
    \begin{center}
        \scalebox{1}[1]{
			\setlength
			\tabcolsep{9.4pt}
			\begin{tabular}{c | c |  c c c | c c c | c  }
                    \toprule
				\multirow{2}{*}{Method} & \multirow{2}{*}{Venue}  & \multicolumn{3}{c|}{RefCOCO} & \multicolumn{3}{c|}{RefCOCO+} & \multicolumn{1}{c}{RefCOCOg} \\ 
                    &  & val & testA & testB & val & testA & testB & val-g \\
                    \hline 

AMR~\cite{AMR} & \textit{AAAI’22}& 14.12          & 11.69          & 17.47                      & 14.13          & 11.47          & 18.13                      & 15.83          \\
GroupViT~\cite{Groupvit} & \textit{CVPR’22}   & 18.03          & 18.13          & 19.33                      & 18.15          & 17.65          & 19.53                      & 19.97          \\

CLIP-ES~\cite{CLIP-ES}& \textit{CVPR’23}                      & 13.79          & 15.23          & 12.87                      & 14.57          & 16.01          & 13.53                      & 14.16          \\

GbS~\cite{GbS}& \textit{ICCV’21}                         & 14.59          & 14.60          & 14.97                      & 14.49          & 14.49          & 15.77                      & 14.21          \\

WWbL~\cite{WWbL}& \textit{NeurIPS’22}                     & 18.26          & 17.37          & 19.90                      & 19.85          & 18.70          & 21.64                      & 21.84          \\

TSEG~\cite{TSEG}& \textit{arXiv’20}                         & 30.12          & -              & -                          & 25.95          & -              & -                          & 22.62          \\

ALBEF~\cite{ALBEF}& \textit{NeurIPS’21}                       & 23.11          & 22.79          & 23.42                      & 22.44          & 22.07          & 22.51                      & 24.18          \\

I-Chunk~\cite{I-Chunk}& \textit{ICCV’23}                       & 31.06          & 32.30          & 30.11                      & 31.28          & 32.11          & 30.13                      & 32.88          \\

TRIS~\cite{TRIS}& \textit{ICCV'23}                                & 31.17          & 32.43          & 29.56                      & 30.90          & 30.42          & 30.80                      & 36.00          \\

APL~\cite{APL} & \textit{ECCV'24}                                           & 55.92          & 54.84          & 55.64                      & 34.92          & 34.87          & 35.61                      & 40.13          \\

QueryMatch~\cite{QueryMatch} & \textit{MM'24}                              & \bm{\textcolor{blue}{59.10}}    & \bm{\textcolor{blue}{59.08}}    & \bm{\textcolor{blue}{58.82}}                & \bm{\textcolor{blue}{39.87}}    & \bm{\textcolor{blue}{41.44}}    & \bm{\textcolor{blue}{37.22}}                & \bm{\textcolor{blue}{43.06}}    \\

Ours          & -                              & 
\bm{\textcolor{red}{61.83}} & 
\bm{\textcolor{red}{62.75}} & 
\bm{\textcolor{red}{60.02}}            & 
\bm{\textcolor{red}{42.05}} & 
\bm{\textcolor{red}{46.39}} &
\bm{\textcolor{red}{37.53}}             & 
\bm{\textcolor{red}{49.06}}\\

                    \bottomrule
			\end{tabular}
		} 
        
	\end{center}
    \vspace{-2mm}
\end{table*}

\begin{table*}[t]
    \caption{Comparisons with state-of-the-art methods on three REC benchmark datasets.}
    \label{tab:rec_results}
    \vspace{-4mm}
    \small
    \begin{center}
        \scalebox{1}[1]{
			\setlength
			\tabcolsep{9.4pt}
			\begin{tabular}{c | c |  c c c | c c c | c  }
                    \toprule
				\multirow{2}{*}{Method} & \multirow{2}{*}{Venue}  & \multicolumn{3}{c|}{RefCOCO} & \multicolumn{3}{c|}{RefCOCO+} & \multicolumn{1}{c}{RefCOCOg} \\ 
                    &  & val & testA & testB & val & testA & testB & val-g \\
                    \hline 

VC~\cite{VC}& \textit{TPAMI'19 }                         & -                    & 32.68                & 27.22                      & -                    & 34.68                & 28.10                       & 29.65                \\
ARN~\cite{ARN}& \textit{ICCV’19 }                            & 32.17                & 35.25                & 30.28                      & 32.78                & 34.35                & 32.13                      & 33.09                \\
KPRN~\cite{KPRN} & \textit{MM'19 }                          & 36.34                & 35.28                & 37.72                      & 37.16                & 36.06                & 39.29                      & 38.37                \\
IGN~\cite{IGN} & \textit{NeurIPS’20 }                         & 34.78                & 37.64                & 32.59                      & 34.29                & 36.91                & 33.56                      & 34.92                \\
DTWREG~\cite{DTWREG}& \textit{TPAMI'21  }                         & 38.35                & 39.51                & 37.01                      & 38.91                & 39.91                & 37.09                      & 42.54                \\

Cycle-Free~\cite{Cycle-Free} & \textit{TMM'21  }                     & 39.58                & 41.46                & 37.96                      & 39.19                & 39.63                & 37.53                      & -                    \\
EARN~\cite{EARN} & \textit{TPAMI'23  }                         & 38.08                & 38.25                & 38.59                      & 37.54                & 37.58                & 37.92                      & 45.33                \\
TGKD~\cite{TGKD} & \textit{ICRA'23 }                           & 39.70                 & 39.92                & 39.63                      & 40.20                 & 39.94                & 40.27                      & 47.99                \\
RefCLIP~\cite{Refclip}& \textit{CVPR'23}                              & 60.36                & 58.58                & 57.13                      & 40.39                & 40.45                & 38.86                      & 47.87                \\
APL~\cite{APL}     & \textit{ECCV'24}                                    & 64.51          & 61.91          & 63.57             & 42.70           &  42.84         &  39.80                 & \bm{\textcolor{blue}{50.22}} \\
QueryMatch~\cite{QueryMatch}& \textit{MM'24 }                                    & \bm{\textcolor{blue}{66.02}}   & \bm{\textcolor{blue}{66.00}}   & \bm{\textcolor{blue}{65.48}}         & \bm{\textcolor{blue}{44.76}}   & \bm{\textcolor{blue}{ 46.72}}   & \bm{\textcolor{blue}{41.50}}              & 48.47       \\



Ours          &-                               & 

\bm{\textcolor{red}{69.03}}       & \bm{\textcolor{red}{70.27}}       & \bm{\textcolor{red}{66.59}}             & \bm{\textcolor{red}{47.16}}      & \bm{\textcolor{red}{52.22}}       & \bm{\textcolor{red}{41.91}}             & \bm{\textcolor{red}{54.72} }           \\

                    \bottomrule
			\end{tabular}
		} 
	\end{center}
    \vspace{-2mm}

\end{table*}

\subsection{Coarse-grained Alignment}
To address category inconsistency, we design a coarse-grained alignment module to first filter out irrelevant visual queries. We discern that the category information is readily available in the input text. For example, it is apparent from the input text \emph{``orange tabby cat standing in a sink''} that the category of the referred object is \emph{``cat''}. We are motivated to predict the specific category and inject this information to ensure that our selected queries belong to the target category. This category-based query-text matching, along with a global query-text matching, effectively mitigates interference from irrelevant objects in the candidate set $\mathcal{Q}_O$. More specifically, at the category matching stage, we inject a Ground Truth (GT) category $c^*$, which is the class label annotation obtained from the dataset. For each query $q_i \in \mathcal{Q}_O$, the Transformer decoder predicts its corresponding category $c^V_i\in\{1, 2, \dots, C\}$ through a classification head, where $C$ is a pre-defined number of total categories (e.g., $C=80$ for MSCOCO \cite{MSCOCO}). The category score measures whether the predicted query category $c^V_i$ is consistent with the GT category $c^*$. If they are the same, the category score $S_{\text{class}, i}$ is set to 1; otherwise, it is set to 0. The above process can be formulated as: 
\begin{equation}
S_{\text{class}, i} = 
\begin{cases} 
1, & \text{if } c^V_i = c^*, \\ 
0, & \text{otherwise}.
\end{cases}
\label{eq:category_matching_score}
\end{equation}

The category-based query-text matching effectively filters out visual queries that belong to irrelevant categories. To fully exploit context in textual representation, we project the query feature $f_{q_i}$ and global text feature $f_t$ into a coarse-grained shared semantic space to learn global visual-linguistic alignment:
\begin{equation}
\bar{f}_{q_i} = {f_{q_i}} \cdotp W_q + b_q,
\label{eq:map_q_i}
\end{equation}
\begin{equation}
\bar{f}_t = f_t\cdotp W_t  + b_t,
\label{eq:map_t}
\end{equation}
where $W_q, W_t$ are projection matrices, $b_q, b_t$ are biases to transform image and text features, respectively. After projection, we calculate the global query-text matching score:
\begin{equation}
S_{\text{global}, i} = \text{sim}(\bar{f}_{q_i}, \bar{f}_t),
\label{eq:feature_similarity_score}
\end{equation}
where $\text{sim}(\cdot)$ is the dot product similarity to measure the alignment between each visual query and the global linguistic feature. 

Overall, we define the coarse-grained alignment score as the weighted sum of the category score and the global score:
\begin{equation}
S_{\text{coarse}, i} = \alpha S_{\text{class}, i} + S_{\text{global}, i},
\label{eq:coarse_grained_similarity_score}
\end{equation}
where $\alpha$ is a hyperparameter to balance the value.

Designed to ensure category consistency, this coarse-grained alignment module filters out category-inconsistent visual queries. In other words, only the queries with $S_{class} = 1$ are selected to construct the refined set $\widetilde{Q}$. We also define a threshold $K$ to curtail the query number based on the coarse-grained score $S_{coarse}$. More details are in the supplemental material.

\subsection{Fine-grained Alignment}
The above coarse-grained alignment module utilizes general linguistic cues to ensure category consistency and filter out category-inconsistent candidates. However, it is insufficient to 
discriminate nuanced semantics and achieve attribute consistency. We further introduce a fine-grained alignment that emphasizes descriptive information in word-level textual features, thereby capturing attribute-aware cross-modal correspondences.

Specifically, we adopt an adaptive phrase attention mechanism \cite{tang2023context} to emphasize linguistic semantics within the word-level features $\mathcal{F}_w$. Instead of focusing on global context or category-level information, this module highlights fine-grained descriptive details by assigning higher attention weights to attribute words and lower weights to category words. For instance, the phrase \emph{``standing in a sink''} provides more discriminative linguistic cues than other words when distinguishing between two cats, and is therefore given greater attention. More precisely, the word-level features $\mathcal{F}_w$ are first processed by a Bidirectional GRU (Bi-GRU) to recalibrate the importance of each word, which can be formulated as:
\begin{equation}
\widetilde{\mathcal{F}}_w = [\overrightarrow{\mathcal{F}}_w, \overleftarrow{\mathcal{F}}_w] = E(\mathcal{F}_w, \theta),
\end{equation}
where $E$ and $\theta$ represent the Bi-GRU module and its parameters, respectively. $\widetilde{\mathcal{F}}_w $ denotes the modulated word-level features that concatenate bidirectional outputs from the Bi-GRU network. To achieve a more adaptive aggregation, we dynamically balance the weights of the predicted features as:
\begin{equation}
\widetilde{\mathcal{F}}_w := \widetilde{\mathcal{F}}_w \cdot \text{softmax}(\text{FC}(\widetilde{\mathcal{F}}_w)),
\end{equation}
where $\text{FC}(\cdot)$ is a fully connected layer to predict the weight assigned to each word.

To learn local visual-linguistic alignment, we aggregate these word-level features and project them into a fine-grained semantically shared space, where the query features are also projected. We formulate the above process as follows:
\begin{equation}
\widetilde{f}_w = f_w\cdot W'_t +b_t', \quad \text{where} \quad f_w=\sum_l \widetilde{\mathcal{F}}_w
\end{equation}
\begin{equation}
\widetilde{f}_{q_i} = {f_{q_i}} \cdotp W_q' + b_q',
\end{equation}
where ${f}_{q_i}$ is the $i$-th query in the refined query set $\widetilde{Q}$. 
The projection matrices are $W_t'$ and $W_q'$, and the bias terms are $b_t' $ and $b_q'$. These parameters are used to transform the image and text features, respectively.


To select the visual query that best matches the text expression at the attribute level, we define the fine-grained alignment score as the dot product similarity between each visual query and the fine-grained adapted text feature, which can be expressed as:
\begin{equation}
S_{\text{fine}, i} = \text{sim}(\widetilde{f}_{q_i}, \widetilde{f}_w),
\end{equation}

Since the adapted word feature $\widetilde{f}_w$ encodes discriminative linguistic semantics, this fine-grained alignment module enables the model to differentiate candidate visual queries based on their local representations, even when they belong to the same category. Finally, we select the query with the highest fine-grained alignment score $S_{\text{fine}, i}$ as the optimal query:
\begin{equation}
{q}^{*} = \underset{i}{\arg\max}\, S_{\text{fine}, i}.
\label{eq:optimal_query_selection}
\end{equation}

\subsection{Training and Inference}
We adopt a query-text contrastive learning strategy \cite{QueryMatch} to achieve weakly supervised learning. A common choice for cross-modal contrastive learning objective is InfoNCE:
\begin{equation}
\mathcal{L}_{cl}(h_t,h_q^+,h_q^-)= -\log\frac{\mathcal{T}(h_t,h_q^+)}{\mathcal{T}(h_t,h_q^+)+\displaystyle\sum_{h_q^-}\mathcal{T}(h_t,h_q^-)},
\end{equation}
where $\mathcal{T} = \exp(\text{sim}(q,k^+)/\tau)$ is the dot product similarity. The text feature $h_t$ should match the visual feature of its designated query $h_q^+$ over a set of negative samples $h_q^-$ from other images. 

In this study, we introduce two shared semantic spaces for visual-linguistic alignment. Therefore, the final contrastive learning objective of our AlignCAT is the sum of that from the two spaces:
\begin{equation}
\mathcal{L}_{CL}= \mathcal{L}_{cl}(\bar{f}_t,\bar{f}_q^+,\bar{f}_q^-) +\mathcal{L}_{cl}(\widetilde{f}_w,\widetilde{f}_q^+,\widetilde{f}_q^-).
\end{equation}

During training, we directly inject the GT category to calculate the category score. However, this information is not available at the inference stage. We are driven to train an auxiliary classifier and predict the category from the text side. Specifically, we add a text classifier to project the global linguistic feature $f_t$ and produce the predicted category, denoted $c^T$. The standard cross-entropy loss is used to train this text classifier:
\begin{equation}
\mathcal{L}_{CE} = - \sum_{i=1}^{C} y_i \log(\hat{y}_i),
\end{equation}
where \( y_i \) is the one-hot encoding of the GT category $c^*$, and \( \hat{y}_i \) is the predicted probability for $i$-th observation belonging to one class.
We note that this text category $c^T$ differs from the query category $c^V_i$ as they are predicted from the linguistic feature $f_t$ and the visual feature $f_{q_i}$, respectively. 

Overall, the weakly supervised learning objective for AlignCAT can be written as follows:
\begin{equation}
\mathcal{L} = \lambda_{1} \mathcal{L}_{CL} + \lambda_{2} \mathcal{L}_{CE},
\label{eq:total_loss}
\end{equation}
where $\lambda_{1}$ and $\lambda_{2}$ are hyperparameters dynamically adjusted to control the strengths, detailed in the supplemental material.

\section{Experiments}
\subsection{Datasets and Metric}
We evaluate the proposed method on three benchmarks: RefCOCO \cite{Refcoco/+},  RefCOCO+ \cite{Refcoco/+}, and RefCOCOg \cite{RefCOCOg}. All of them are based on MSCOCO \cite{MSCOCO}, and each contains (image, expression) as: (19,994, 142,210), (19,992, 141,564), (26,711, 104,560). In these three datasets, each expression is associated with one class label, which is used as the GT category in the coarse-grained alignment. Regarding the text expression, RefCOCO describes objects with absolute spatial information, while the other two datasets are more challenging. RefCOCO+ focuses more on relative spatial information and appearance (such as color and texture), and RefCOCOg provides longer expressions that are more complex and carry richer semantics. For the REC task, we follow \cite{Refclip,QueryMatch} that use IoU@0.5 as the metric. We count a prediction as correct if the IoU between the predicted and GT bounding boxes exceeds 0.5. For the RES task, we adopt mIoU \cite{CLIP-ES,WWbL} as the metric that calculates the average IoU across all test samples. More details are in the supplementary material.

\subsection{Implementation Details}
Following \cite{QueryMatch}, we employ the pretrained Mask2Former detector \cite{mask2former} and freeze its parameters when training our AlignCAT. The image resolution is set to $416\times 416$. The text lengths for RefCOCO, RefCOCO+, and RefCOCOg are 15, 15, and 20, respectively. All experiments are conducted on two 24G Nvidia RTX 4090 GPUs. The batch size per GPU is 14. The query feature dimension is 256, and the dimensions for word-level features, text features, and the shared semantic space are all 512. During query selection, we set $O=20$ for confidence-based filtering, $K=10$ for the maximum selected queries after coarse-grained alignment. We set $\alpha=100$ to emphasize the category information for calculating coarse-grained scores. 
We use the Adam optimizer \cite{adam} with a learning rate of $1e-4$ and set training epochs to 25.

\begin{table}
    \caption{Ablation of the formula for query quality estimation.}
    \vspace{-3mm}
    \label{tab:Components}
	\centering
			\begin{tabular}{c|ccc}
				\toprule
                Formula & val & testA  & testB \\
				\hline
                       $S_{\text{global}}$&  65.89                   & 65.94                     & 65.47 \\
                    $S_{\text{global}}+S_{\text{class}}$& 67.55$_{\uparrow 1.66}$                   & 69.63$_{\uparrow 3.69}$                       & 64.66$_{\downarrow 0.81}$  \\

                      $S_{\text{global}}+S_{\text{fine}}$&
                      
                      67.21$_{\uparrow 1.32}$                    & 67.93$_{\uparrow 1.99}$                      & 66.21$_{\uparrow 0.74}$                      \\

                      $S_{\text{fine}}+S_{\text{global}}+S_{\text{class}}$& 
                      67.36$_{\uparrow 1.47}$           & 68.66$_{\uparrow 2.72}$              & 66.48$_{\uparrow 1.01}$ \\              \textbf{$S_{\text{global}}+S_{\text{class}}+S_{\text{fine}}$} & 
                      \textbf{69.03$_{\uparrow 3.14}$} & \textbf{70.27$_{\uparrow 4.33}$} & \textbf{66.59$_{\uparrow 1.12}$}
                      \\ 
				\bottomrule
			\end{tabular}
    \vspace{-1mm}
\end{table}

\begin{table}
    \caption{Ablation study of the injected category information. ``Train'' and ``Infer'' refer to the training and inference stages. $c^*$: GT category. $c^T$: text classifier's predicted category. }
    \vspace{-3mm}
    \label{tab:cat_utilization}
    \centering
    \small
    
    \renewcommand\arraystretch{1}
    \scalebox{0.95}{
        \setlength\tabcolsep{5pt}
        \begin{tabular}{cc|c|ccc}
            \toprule
            $c^*$ \footnotesize{(\textit{Train})} & $c^{T}$ \footnotesize{(\textit{Train})} & $c^{T}$ \footnotesize{(\textit{Infer})}  & val  & testA &testB  \\
            \hline
            -&- &  -& 67.21   & 67.93  & 66.21  \\
            
            $\checkmark$ & -&- &   68.74$_{\uparrow 1.53}$   & 
            69.84$_{\uparrow 1.91}$  & 66.23$_{\uparrow 0.02}$   \\

            -& - & $\checkmark$  & 67.61$_{\uparrow 0.40}$   & 68.18$_{\uparrow 0.25}$  & 66.40$_{\uparrow 0.19}$  \\

            -& $\checkmark$ & $\checkmark$ & 64.64$_{\downarrow 2.57}$  & 64.90$_{\downarrow 3.03}$  & 63.53$_{\downarrow 2.68}$ \\

            $\bm{\checkmark}$ & \textbf{-} &  $\bm{\checkmark}$  & \textbf{69.03$_{\uparrow 1.82}$} & \textbf{70.27$_{\uparrow 2.34}$} & \textbf{66.59$_{\uparrow 0.38}$} \\
            \bottomrule
        \end{tabular}}
    \vspace{-2mm}
\end{table}

\subsection{Quantitative Analysis}
In this section, we first validate AlignCAT by comparing it with comprehensive weakly supervised VG methods, and ablate key components of our approach.

\textbf{Comparison to the state-of-the-arts}. In Tables \ref{tab:res_results} and \ref{tab:rec_results}, we compare AlignCAT with a set of weakly supervised VG methods. The first observation is that AlignCAT significantly outperforms existing methods on all three benchmarks. Our method improves the average accuracy by $+2.53\%$ and $+2.80\%$ over QueryMatch on RefCOCO for RES and REC, respectively. The improvement on RefCOCOg is particularly notable, with AlignCAT increasing the accuracy of QueryMatch by more than $6\%$ for both tasks. We also notice that AlignCAT excels on TestA of all datasets, where most categories of referred objects are \emph{``person''}. With the help of category matching, AlignCAT effectively filters out visual queries not belonging to humans, before more fine-grained alignment. This validates the effectiveness of our innovative category-then-attribute mechanism in enhancing cross-modal alignment, with the capacity to tackle multi-object images and complex text expressions.


\begin{table}
    \caption{Ablation of the module to inject GT category.}
    \vspace{-4mm}
    \label{tab:gt_cinf}
	\centering
	\small
	
	\renewcommand\arraystretch{1}
	\begin{center}
		\scalebox{1}[1]{
			\setlength
			\tabcolsep{7.2pt}
			\begin{tabular}{cc|ccc}
				\toprule
				Confidence-based    & Coarse-grained  & \multirow{2}{*}{val} & \multirow{2}{*}{testA}  & \multirow{2}{*}{testB} \\
                Selection   & Alignment \\

				\hline
                -& -& 67.61 & 68.18 & 66.40 \\
                $\checkmark$ &- &  \multicolumn{1}{l}
                {67.24} & \textbf{71.50}        & 62.06    \\
                
                - & $\bm{\checkmark}$ &\textbf{69.03} & 70.27                      & \textbf{66.59}           \\		
				\bottomrule
			\end{tabular}
		}
	\end{center}
	\vspace{-2mm}
\end{table}

\begin{figure}[t!]
	\includegraphics[width=0.92\linewidth]{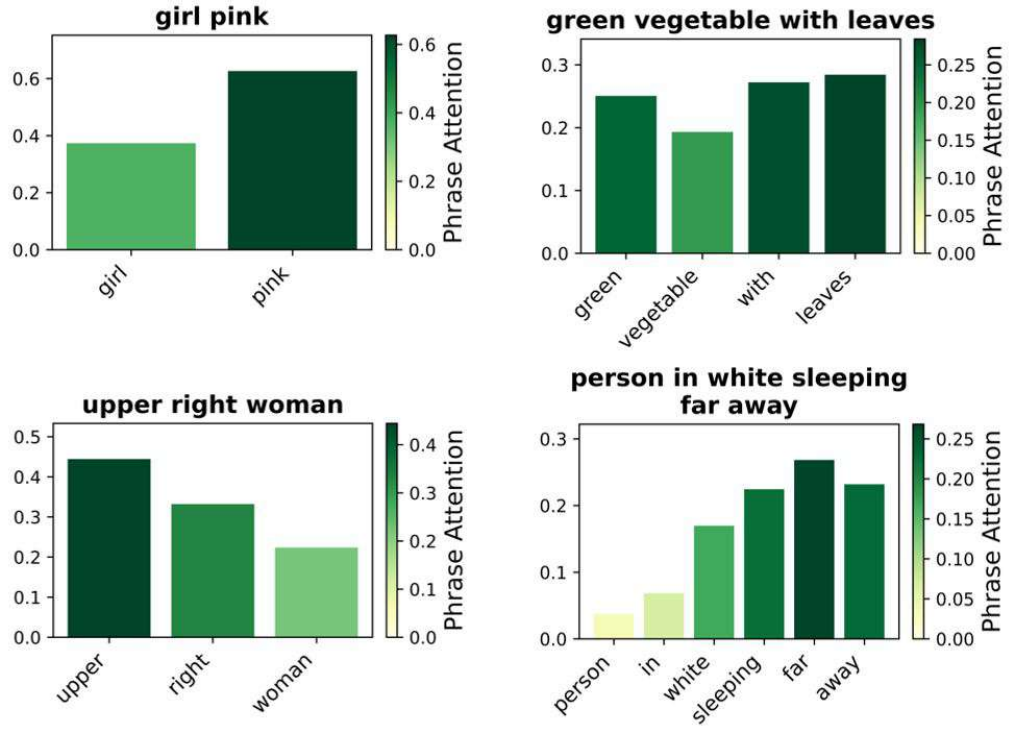} %
    \vspace{-3mm}
	\caption{Visualization of adaptive phrase attention.}
    \vspace{-3mm}
	\label{fig:atten}
\end{figure}
\begin{figure*}[t!]
	\centering
	\includegraphics[width=0.95\textwidth]{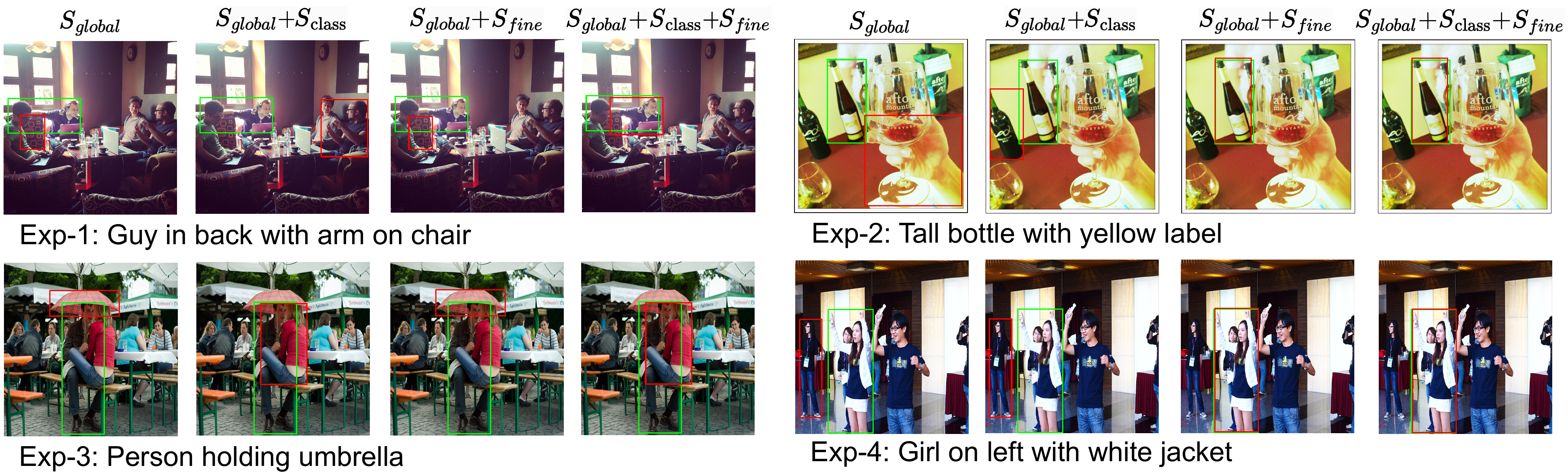} %
    \vspace{-0.3cm}
	\caption{Visualization comparison of different selection designs of AlignCAT in weakly supervised REC. The red and green boxes are GT and predicted grounding results, respectively.}
	\label{fig:DifferentDesigns}
    
\end{figure*} 

\begin{figure*}[t!]
	\centering

	\includegraphics[width=0.95\textwidth]{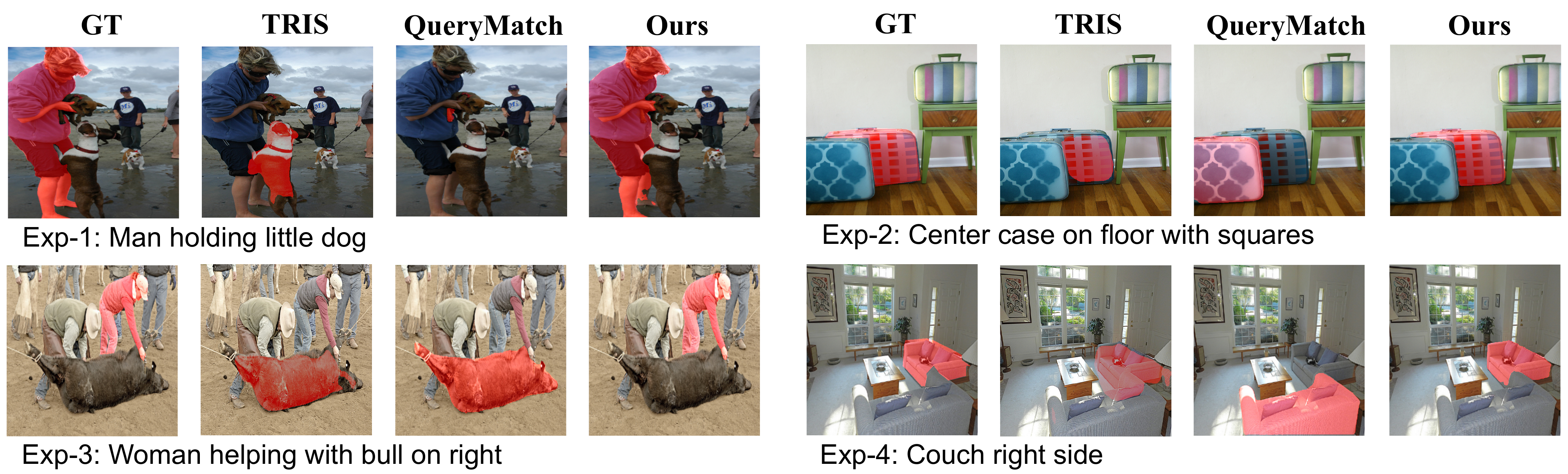} %
    \vspace{-0.3cm}
	\caption{Visualization comparison to TRIS and QueryMatch for the weakly supervised RES task. The GT and predicted segmentation results are marked in red.}
	\label{fig:fig_resresults}
    
\end{figure*}


\textbf{Ablation of AlignCAT}. 
To validate the designs of AlignCAT, we have conducted various ablation studies on the RefCOCO dataset for weakly supervised REC.
We first compare different settings of query selection. When ablating the design of global similarity, the corresponding contrastive learning objective is also removed. The same applies to the fine-grained alignment with word-level similarity calculation.
These results are reported in Table \ref{tab:Components}. The baseline selects one positive visual query with the highest $S_{\text{global}}$. With category matching, the combination $S_{\text{class}} + S_{\text{global}}$ improves VG performance on two subsets, albeit with a slight decrease on testB. This suggests that category information benefits human-target localization, but struggles with non-human objects. Solely using the fine-grained alignment, $S_{\text{global}} + S_{\text{fine}}$ achieves $67.93\%$ on RefCOCO testA, yet is worse than the former setting with $69.63\%$. This comparison highlights the importance of category-based filtering. We also experimented with the attribute-then-category order. The result of $S_{\text{fine}} + S_{\text{global}} + S_{\text{class}}$ shows a remarkable performance decline compared to $S_{\text{global}} + S_{\text{class}}$. We suspect that without category-based filtering, the text features of contextual objects create noise and affect the cross-modal alignment. Conversely, $S_{\text{class}} + S_{\text{global}} + S_{\text{fine}}$ with a category-then-attribute order achieves the best performance, demonstrating the effectiveness of the coarse-to-fine visual-linguistic matching scheme, as well as the complementary effect of three selection modules. 

Next, we examine different strategies for using category information during training and inference. As shown in Table \ref{tab:cat_utilization}, the first row is the baseline without category information. The second row is the model trained with the GT category $c^*$ while removing the category matching score during inference. This setting improves the model performance, which highlights the importance of category information in enhancing cross-modal correspondences. The third row is the result of training the text classifier but only injecting the predicted category $c^T$ during inference, which presents a slight improvement. Notably, in the last second row, directly injecting the predicted category $c^T$ during training indicates a significant performance decline. We suspect that the text classifier's predicted categories are largely inaccurate at the beginning of training, resulting in unreliable visual-linguistic alignment. The last row is our full model that uses the GT category during training and injects the predicted category during inference. This design enhances robustness and achieves the best performance across all subsets. 

We further investigate alternative strategies for injecting GT category information. As shown in Table \ref{tab:gt_cinf}, we compare the confidence-based selection and the global feature alignment to incorporate the category score. In the former, $\mathcal{Q}_O$ is filtered by confidence and category matching, while $\widetilde{Q}$ relies on global similarity $S_{\text{global}}$.
Although this improves testA performance, 
it leads to a significant performance degradation on testB. This issue arises due to suboptimal negative query selection, which are sampled from $\mathcal{Q}_O$. Since a large proportion of referred objects in the training set belong to the \emph{``person''} category, integrating category matching during confidence-based filtering results in the same category between most negative queries and the positive query. This reduces the diversity of negative samples and affects generalization. To address this, we inject the category information at the coarse-grained alignment module. This setting enhances negative sample quality and improves the model's robustness across comprehensive scenarios.

\begin{figure*}[t!]
	\centering
	\includegraphics[width=0.98\textwidth]{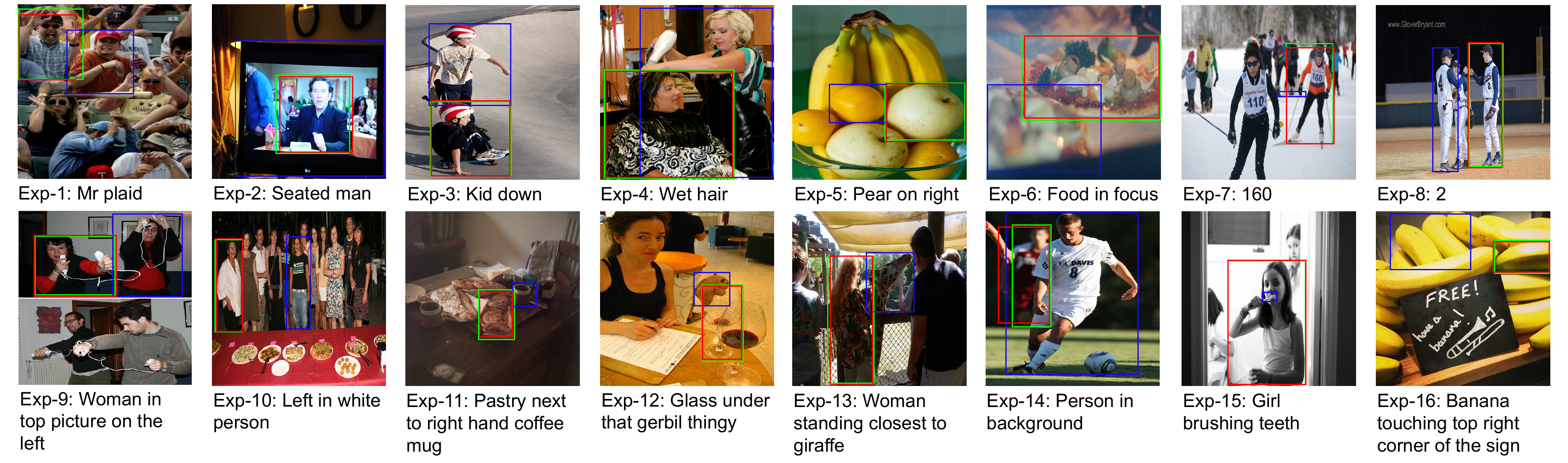} %
    \vspace{-4mm}
	\caption{Visualization comparison in weakly supervised REC. 
    \textcolor{green}{Green: ground truth}. \textcolor{blue}{Blue: QueryMatch}. \textcolor{red}{Red: Ours}.}
	\label{fig:rec_test}
    \vspace{-1mm}
\end{figure*} 
\begin{figure}[t!]
	\centering
	\includegraphics[width=1\linewidth]{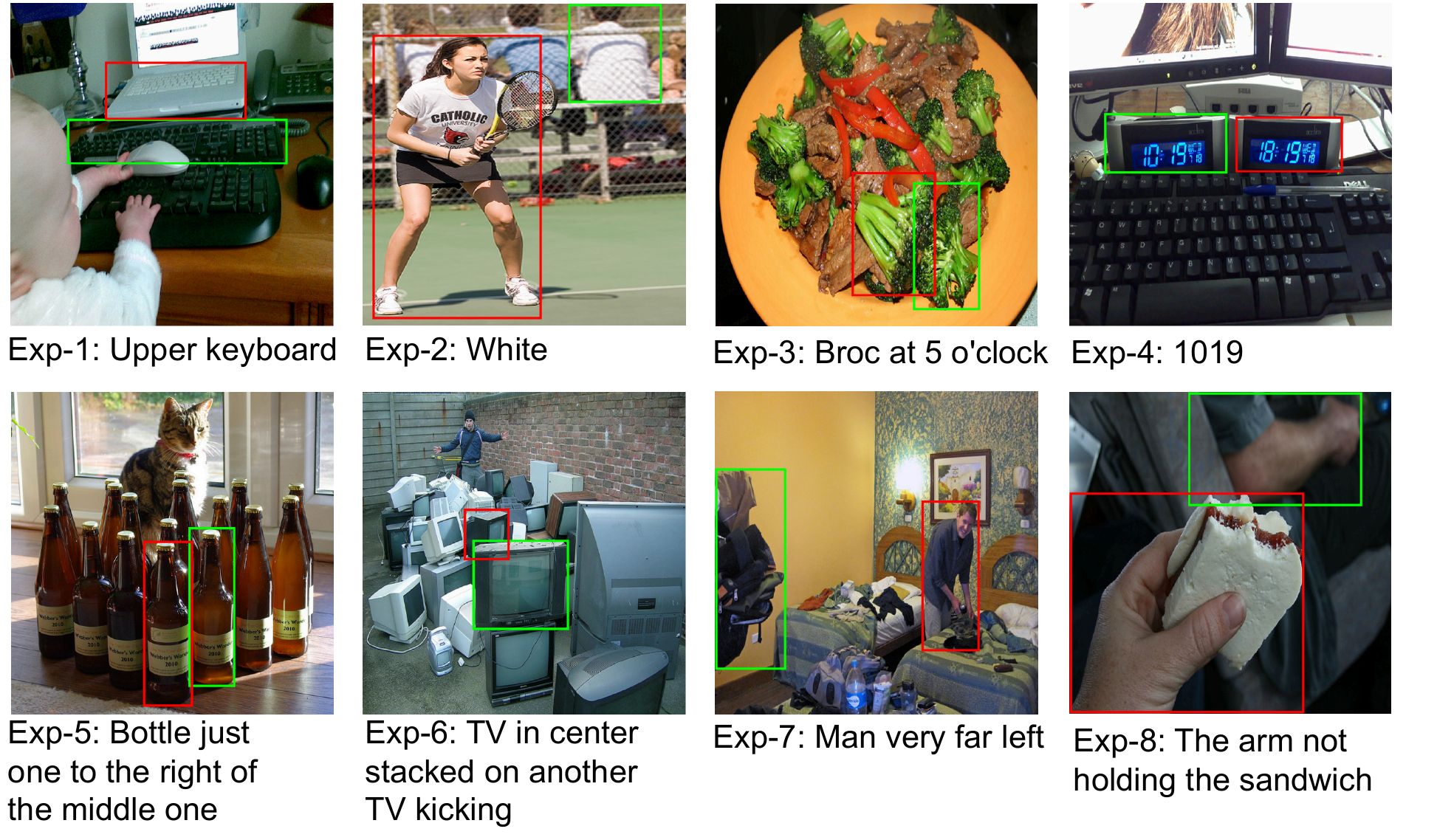} %
    \vspace{-6mm}
	\caption{Failure cases of AlignCAT in weakly supervised REC. 
    \textcolor{green}{Green: ground truth}. \textcolor{red}{Red: Ours}.}
	\label{fig:fail_case}
    \vspace{-3mm}
\end{figure} 
\subsection{Qualitative Analysis}
In Figure \ref{fig:atten}, we visualize the weights of the modulated word-level text features after adaptive phrase attention. These values illustrate how the model dynamically adjusts the importance of each word. For example, given the text \emph{``girl pink''}, the model highlights the color attribute \emph{``pink''} than the category word \emph{``girl''}. Interestingly, the contextual object \emph{``leaves''} is allocated with a higher value than the referred object \emph{``vegetable''}. This observation explains the result in Table \ref{tab:Components} for the inferior performance in the setting of the exchanged order. Overall, AlignCAT leverages descriptive information to mitigate intra-class ambiguity, thereby distinguishing objects belonging to the same category.


To gain in-depth insights into the category-then-attribute alignment mechanism, we ablate AlignCAT with four configurations and visualize the results in Figure \ref{fig:DifferentDesigns}. Utilizing solely the global feature similarity $S_{\text{global}}$ struggles to achieve category and attribute consistencies, especially when images involve multiple objects. With the category matching score, the results present the category consistency. For instance, $S_{\text{global}}+S_{\text{class}}$ excludes the contextual object \emph{``label''} in Exp-2, however, it fails to resolve intra-class ambiguities and selects another bottle. On the other hand, without $S_{\text{class}}$, the results of $S_{\text{global}} + \text{S}_{\text{fine}}$ still suffers from category inconsistency in Exp-3, which highlights the importance of the category matching. This discovery is consistent with the quantitative comparison in Table \ref{tab:Components}. In contrast, the full configuration $S_{\text{global}} + S_{\text{class}} + \text{S}_{\text{fine}}$ achieves category and attribute consistencies, whether the target is human or non-human, and the attribute is color (e.g., \emph{``white''}) or spatial relation (e.g., \emph{``in back''}). Notice that AlignCAT may fail to accurately locate the target object due to the occlusion problem (e.g., the arm behind the head in Exp-1).

In Figure \ref{fig:fig_resresults}, we compare AlignCAT with two state-of-the-art weakly supervised RES models, TRIS \cite{TRIS} and QueryMatch \cite{QueryMatch}. Given texts with complex relationships and intensive images with multiple objects, they incorrectly locate the contextual objects such as \emph{``dog''} and \emph{``bull''}. Conversely, the proposed method achieves higher segmentation accuracy. With stronger reasoning ability and better visual understanding, AlignCAT is more robust and reliable in extensive grounding scenarios.

Figure \ref{fig:rec_test} compares the performance of AlignCAT and QueryMatch in the weakly supervised REC task. The analysis shows that AlignCAT has a clear advantage in maintaining category and attribute consistency. For example, in Exp-7, AlignCAT successfully aligns the abstract and vague query \emph{``160''} with the \emph{``person''} category, accurately localizing the target. In contrast, QueryMatch fails to understand this abstract query, leading to a localization failure. In complex multi-object scenarios, AlignCAT continues to effectively select the correct queries. For instance, in Exp-16, AlignCAT not only identifies the logical subject \emph{``banana''}, but also captures its fine-grained spatial relationships, achieving precise localization. In summary, AlignCAT, with its robust coarse-to-fine semantic alignment, outperforms QueryMatch in complex scenarios.

To provide a more comprehensive analysis, we present typical failure cases of AlignCAT for the REC task in Figure \ref{fig:fail_case}. These failures occur due to dataset quality issues, including wrong ground truth annotations (Exp-1) and insufficient textual descriptions (Exp-2). 
Meanwhile, our method still lacks semantic understanding to comprehend out-of-distribution textual expressions (Exp-3), or to discern nuanced visual features for similar objects (Exp-4). 


\section{Conclusion}
In this study, we identify that existing weakly supervised VG methods suffer from contextual ambiguities, showing category and attribute inconsistencies. To address these challenges, we propose a novel query-based VG framework, AlignCAT, with a category-then-attribute visual-linguistic alignment strategy to progressively filter out query candidates. To ensure category consistency, we design a coarse-grained alignment module that leverages category information and global context. For attribute consistency, we further propose a fine-grained alignment module to capture word-level linguistic features and emphasize attribute-based query-text alignment, effectively resolving intra-class ambiguities. Extensive experiments demonstrate that AlignCAT achieves state-of-the-art performance on three benchmarks for both REC and RES tasks. The proposed category-then-attribute alignment enhances category and attribute consistencies in comprehensive scenes. This work provides novel insights into leveraging linguistic cues for advancing weakly supervised visual grounding.

\begin{acks}
This work was partially supported by the National Natural Science Foundation of China (No. 62272227). It was also partly supported by the MUR PNRR project FAIR (PE00000013) funded by the NextGenerationEU, the EU Horizon projects ELIAS (No. 101120237) and ELLIOT (No. 101214398).
\end{acks}

\bibliographystyle{ACM-Reference-Format}
\bibliography{sample-base}


\clearpage

\appendix

\setcounter{table}{5}   
\setcounter{figure}{7}

\section{Details of the proposed method}

\subsection{Details of coarse-grained selection quantity}
To construct the refined set $\widetilde{Q}$ in the coarse-grained alignment, we develop an adaptive strategy. Let $M$ represent the number of visual queries where $S_\text{class} = 1$. We define a threshold $K$ as the maximum number of queries that can be selected(default to 10). If $0<M<K$, all queries belonging to the target category are selected. If $M\geq K$ or $M=0$, we keep the top-$K$ queries ranked in descending order by the score $S_\text{coarse}$. 

\subsection{Details of Dynamic Training}

In Equation (17), we introduce two hyperparameters, $\lambda_{{1}}$ and $\lambda_{{2}}$, to dynamically adjust the weakly supervised learning objective. These hyperparameters aim to balance the learning focus of different tasks during training. Specifically, at the start of training, we assign a higher weight to $\lambda_{2}$. This helps the model quickly learn to predict text categories correctly. As training progresses, we gradually reduce the value of $\lambda_{2}$ using an exponential decay strategy. This shift directs the model's attention toward contrastive learning tasks. This gradual adjustment balances category prediction and contrastive learning objectives throughout training, enhancing the model's stability and generalization ability.

\section{Ablation Details of Category Information Utilization Strategy}
In the main paper, we ablate five configurations of how to use categorical information in Table 4. 
Specifically, referring to the first row, does not use category information and relies solely on $S_{\text{global}}+S_{\text{fine}}$ for positive sample selection, where \(S_{\text{global}}\) represents the global similarity and \(S_{\text{fine}}\) represents the fine-grained alignment score.
In the second row, When we introduced GT category information for coarse-grained alignment during training, the positive sample selection strategy became $S_{\text{global}} + S_{\text{class}} + S_{\text{fine}}$ during training and $S_{\text{global}} + S_{\text{fine}}$ during inference. Here,  $S_{\text{class}}$ represents the category score for coarse alignment using GT category information. This configuration improved the model's training quality and overall performance by leveraging GT category information.
In the third row, we experimented with not using GT category information during training but employing text classifier-predicted categories for coarse-grained alignment during inference. In this setup, the positive sample selection strategy was $S_{\text{global}} + S_{\text{fine}}$ during training and $S_{\text{global}} + \widehat{S}_{class} + S_{\text{fine}} $ during inference, where $\widehat{S}_{class}$ represents the  category score for coarse alignment using predicted categories. In the fourth row, We tested a configuration entirely dependent on text classifier-predicted categories, using $S_{\text{global}} + \widehat{S}_{class} + S_{\text{fine}}$ for both training and inference. 
In the last row, the best-performing configuration combined GT category information during training and text classifier-predicted categories during inference, using $S_{\text{global}} + S_{\text{class}} + S_{\text{fine}}$ for training and $ S_{\text{global}} + \widehat{S}_{class} + S_{\text{fine}}$ for inference. This final setup effectively combined the benefits of GT category information in training with the practicality of text classifiers in inference, achieving the best performance across all configurations.

\begin{figure}[t!]
	\includegraphics[width=\linewidth]{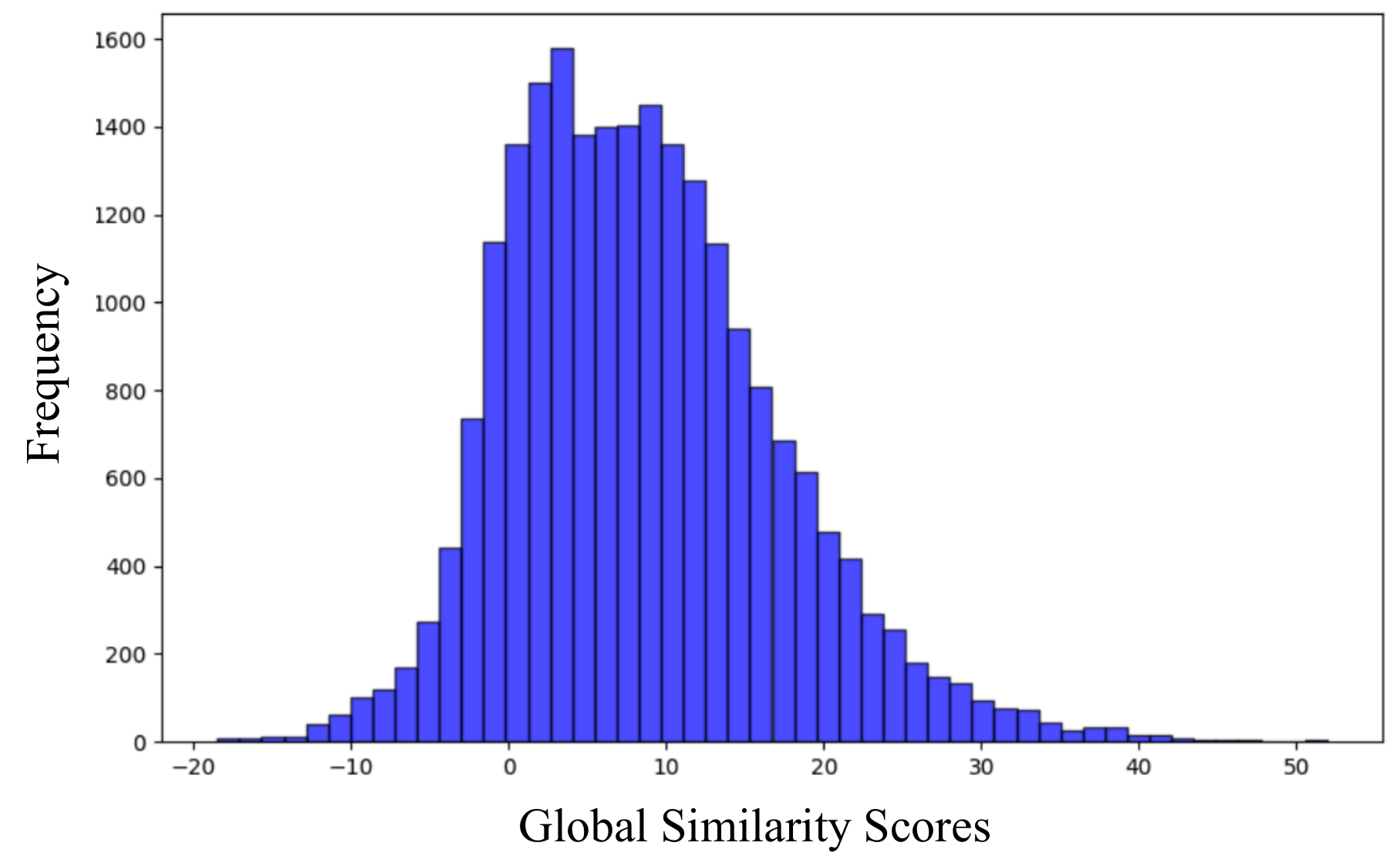} %
	\caption{Visualization of global similarity scores $S_{global}$.}
	\label{fig:histogram}
\end{figure}

\section{Ablation of $\alpha$ in Coarse-Grained Alignment}
\begin{table}

	\centering
	\small
	
	\renewcommand\arraystretch{1}
	\begin{center}
		\scalebox{1}[1]{
			\setlength
			\tabcolsep{7.2pt}
			\begin{tabular}{c|ccc}
				\toprule
                \multirow{2}{*}{$\alpha$}

				 & \multicolumn{3}{c}{Accuracy(\%)}  \\
                 
				 &  val & testA  & testB \\ \hline
			
                    0 &   66.25& 66.38 &   65.54 \\ 
                    1 &   67.17& 67.80 &   65.73 \\ 
                    10 & 68.01  & 68.23  & 65.79   \\
                    20 & 68.45  & 68.96 &  65.89  \\
                    100 & \textbf{68.46}  & \textbf{68.96} &  \textbf{65.91}  \\
			
				\bottomrule
			\end{tabular}
		}
        \caption{Ablation of $\alpha$ in coarse-grained alignment score for Weakly Supervised REC on RefCOCO.}
        \label{tab:alpha_coarse} 
    \end{center}
    
	\vspace{-0.4cm}

\end{table}
We introduce a hyperparameter \(\alpha\) in Equation (9) to control the weight of \(S_{\text{coarse}}\). This helps balance it with the global score \(S_{\text{global}}\) in the coarse alignment module.
To adjust $\alpha$ appropriately, we investigate the distribution of the global similarity score $S_{\text{global}}$. As shown in Figure \ref{fig:histogram}, we randomly selected 30,960 query-text pairs and calculated their global similarity scores. Observe that the plot follows a typical bell-shaped distribution, mainly concentrated on $[-3,28]$. Based on this discovery, we set 5 different values of  $\alpha$ and analyzed their effect. As listed in Table \ref{tab:alpha_coarse} as $\alpha$ increases, the REC accuracy gradually improves. When $\alpha=0$, the coarse-grained alignment score $S_\text{coarse}$ is solely determined by $S_\text{global}$. Notice that even a tiny ratio of the category score, i.e., $\alpha=1$, leads to notable improvement ($\sim1.5$ on testA). This comparison highlights the crucial role of category information in coarse-grained alignment. When $\alpha=100$, the weight of the category score is dominant and significantly improves the model performance. By increasing the weight of category scores, the model accurately filters out non-target categories, improving the quality of the candidate object set for more reliable input to the subsequent fine-grained alignment module. Ultimately, this ablation validates the importance of category information in enhancing the model's overall performance.

\begin{table}
\caption{Text expression examples and their corresponding categories in the dataset.}
	\centering
	\small
	
	\renewcommand\arraystretch{1}
	\begin{center}
		\scalebox{1}[1]{
			\setlength
			\tabcolsep{7.2pt}
			\begin{tabular}{c|c}
				\toprule
                Category & Text\\ \hline
                \multirow{5}{*}{Person}
                     &  \emph{``2''}\\ 
                     &  \emph{``left kid''}\\
                     &  \emph{``upper right corner''}\\
                     &  \emph{``kid looking at you''}\\
                     &  \emph{``the man with an umbrella''}\\ \hline
                \multirow{5}{*}{Car}
                     &  \emph{``wheel''}\\ 
                     &  \emph{``front taxi''}\\
                     &  \emph{``bottom left car''}\\
                     &  \emph{``partial white car nearest us''}\\ 
                     &  \emph{``blue car behind red shirt kid''}\\ 
                     \hline

                \multirow{5}{*}{Suitcase}
                     &  \emph{``blue''}\\ 
                     &  \emph{``luggage left''}\\
                     &  \emph{``right box on cart''}\\ 
                    &  \emph{``not those red luggage''}\\
                     &  \emph{``red suitcase in middle of the 3''}\\
                     \hline  

                \multirow{5}{*}{Cat}
                     &  \emph{``top''}\\ 
                     &  \emph{``bggest cat left''}\\
                    &  \emph{``cat on topbehind''}\\
                     &  \emph{``bottom pic leftmost animal''}\\ 
                     &  \emph{``kitty at bottom of picture''}\\

				\bottomrule
			\end{tabular}
		}
        
        \label{tab:category_text}
	\end{center}
	\vspace{-0.4cm}

\end{table}


\section{Display of Dataset Text and Corresponding Categories}
In table \ref{tab:category_text}, we present visualizations of the text expressions in the dataset along with their corresponding categories. It is clear that, although some text expressions (e.g., \emph{``upper right corner"}) alone may not clearly refer to an object, incorporating the relevant category information (e.g., \emph{``person"}) helps narrow down the location and improves the model's ability to accurately identify the target. On the other hand, abstract descriptions like \emph{``2"} or \emph{``blue"} may lack sufficient location details, but when combined with category information, they help the model better define the target category. In multi-object scenarios, category information significantly enhances the clarity of text references, especially in scenes with multiple object categories. It provides strong support for the model and helps resolve category inconsistency effectively.

\section{Reducing Reliance on Ground-Truth Labels for Category Prediction}
To improve generalizability and reduce reliance on ground-truth labels during training, and reduce reliance on ground-truth category labels during training, one viable approach is to pre-train this text classifier. We report the text classification accuracy of the pre-trained text classifier on RefCOCO+ as 89.5\% (val), 96.8\% (testA), and 78.6\% (testB). We also explore LLMs to predict category labels. Specifically, we prompt GPT-4o to select one category from COCO 80 categories for each input expression. Since each sample is paired with multiple expressions, we consider two settings: (1) using a randomly selected expression; (2) using all available expressions (similar to [7]). The performance of GPT-4o on 1,000 samples from each of the three training datasets is summarized in Table \ref{tab:gpt_predict}.
We observe that GPT-4o achieves significantly higher accuracy when multiple expressions are used. These results also highlight issues with ambiguous or low-quality annotation, e.g., in Fig. 7 Exp-2, the expression "White" refers to a person, making it inherently difficult to classify the label correctly.
We believe incorporating more robust category prediction strategies, e.g., classifier pre-training, LLM guidance, and annotation preprocessing, can enhance the generalizability of AlignCAT in real-world applications. We plan to investigate this direction thoroughly in future work.

\begin{table}

	\centering
	\small
	
	\renewcommand\arraystretch{1}
	\begin{center}
		\scalebox{1}[1]{
			\setlength
			\tabcolsep{7.2pt}
			\begin{tabular}{c|ccc}
				\toprule
                \multirow{2}{*}{Setting}

				 & \multicolumn{3}{c}{Accuracy(\%)}  \\
                 
				 &  RefCOCO & RefCOCO+  & RefCOCOg \\ \hline
			
                    Single expressions &   86.2& 79.9 &   90.3 \\ 
                    Multiple expressions &   95.8 & 96.1 &   97.5 \\ 
			
				\bottomrule
			\end{tabular}
		}
        \caption{Accuracy of Category Prediction Using GPT-4o}
        \label{tab:gpt_predict} 
    \end{center}
    \vspace{-0.4cm}

\end{table}

\end{document}